\documentclass{article}

\PassOptionsToPackage{numbers, compress}{natbib}

\usepackage[final]{neurips_2025}

\usepackage[utf8]{inputenc} 
\usepackage[T1]{fontenc}    
\usepackage{hyperref}
\usepackage{url}            
\usepackage{booktabs}       
\usepackage{amsfonts}       
\usepackage{nicefrac}       
\usepackage{microtype}      
\usepackage[table]{xcolor}         
\usepackage{amsmath}
\usepackage{array}     
\usepackage{graphicx}
\usepackage{multirow} 
\usepackage{float}
\usepackage{subcaption}
\usepackage{textcomp}

\title{DreamPRM: Domain-Reweighted Process Reward Model for Multimodal Reasoning}

\author{%
  Qi Cao \\
  University of California, San Diego\\
  \texttt{q9cao@ucsd.edu} \\
  \And
  Ruiyi Wang \\
  University of California, San Diego\\
  \texttt{ruiyi@ucsd.edu} \\
  \And
  Ruiyi Zhang \\
  University of California, San Diego\\
  \texttt{ruz048@ucsd.edu} \\
  \And
  Sai Ashish Somayajula \\
  University of California, San Diego\\
  \texttt{ssomayaj@ucsd.edu} \\
  \And
  Pengtao Xie \\
  University of California, San Diego\\
  Mohamed bin Zayed University of Artificial Intelligence (MBZUAI)\\
  \texttt{p1xie@ucsd.edu} \\
}

\begin{document}

\maketitle
\vspace{-1em}
\begin{center}
    Project Page: \href{https://github.com/coder-qicao/DreamPRM}{\texttt{https://github.com/coder-qicao/DreamPRM}}
\end{center}

\begin{abstract}
Reasoning has substantially improved the performance of large language models (LLMs) on complicated tasks. Central to the current reasoning studies, Process Reward Models (PRMs) offer a fine-grained evaluation of intermediate reasoning steps and guide the reasoning process. However, extending PRMs to multimodal large language models (MLLMs) introduces challenges. Since multimodal reasoning covers a wider range of tasks compared to text-only scenarios, the resulting distribution shift from the training to testing sets is more severe, leading to greater generalization difficulty. Training a reliable multimodal PRM, therefore, demands large and diverse datasets to ensure sufficient coverage. However, current multimodal reasoning datasets suffer from a marked quality imbalance, which degrades PRM performance and highlights the need for an effective data selection strategy. To address the issues, we introduce DreamPRM, a domain-reweighted training framework for multimodal PRMs which employs bi-level optimization. In the lower-level optimization, DreamPRM performs fine-tuning on multiple datasets with domain weights, allowing the PRM to prioritize high-quality reasoning signals and alleviating the impact of dataset quality imbalance. In the upper-level optimization, the PRM is evaluated on a separate meta-learning dataset; this feedback updates the domain weights through an aggregation loss function, thereby improving the generalization capability of trained PRM. Extensive experiments on multiple multimodal reasoning benchmarks covering both mathematical and general reasoning show that test-time scaling with DreamPRM consistently improves the performance of state-of-the-art MLLMs. Further comparisons reveal that DreamPRM's domain-reweighting strategy surpasses other data selection methods and yields higher accuracy gains than existing test-time scaling approaches. Notably, DreamPRM achieves a top-1 accuracy of 85.2\% on the \textsc{MathVista} leaderboard using the o4-mini model, demonstrating its strong generalization in complex multimodal reasoning tasks.
 
\end{abstract}

\section{Introduction}
\begin{figure}[t]
  \centering
  \includegraphics[width=\textwidth]{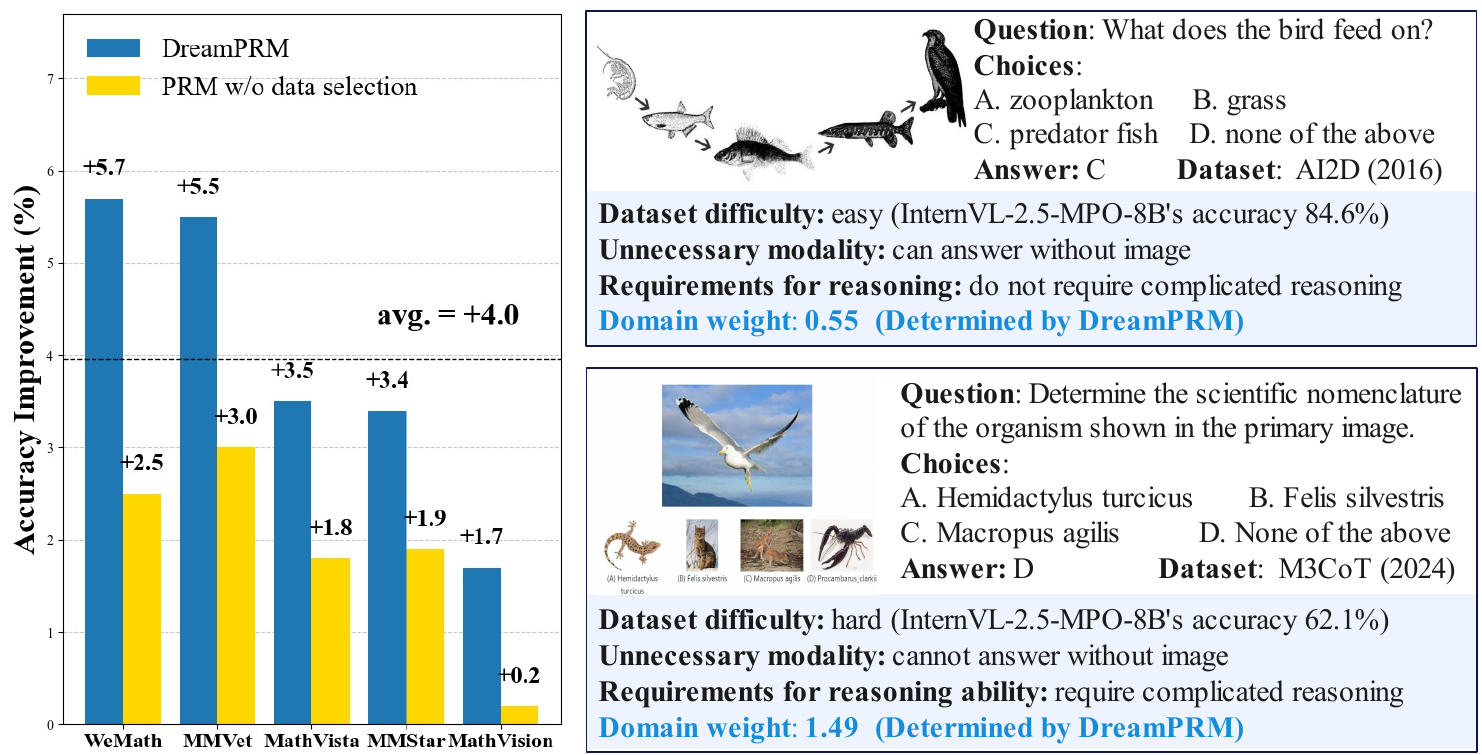} 
  \caption{\textbf{DreamPRM improves multimodal reasoning by mitigating the dataset quality imbalance problem.} \textbf{Left}: On five benchmarks, DreamPRM outperforms base model (InternVL-2.5-8B-MPO~\citep{wang2024mpo}) by an average of \(+4.0\%\). DreamPRM also consistently surpasses Vanilla PRM trained without data selection. \textbf{Right}: Easy \textsc{AI2D}~\citep{kembhavi2016diagramworthdozenimages} questions (weight 0.55) vs.\ hard \textsc{M3CoT}~\citep{chen2024m3cotnovelbenchmarkmultidomain} questions (weight 1.49) shows how DreamPRM prioritizes data that demand deeper reasoning -  samples requiring knowledge from both textual and visual modalities for step-by-step logical deduction.}

\label{fig:intro}
\end{figure}
Reasoning~\citep{snell2024scalingllmtesttimecompute} has significantly enhanced the logical and critical thinking capabilities of large language models (LLMs)~\citep{brown2020languagemodelsfewshotlearners, chowdhery2022palmscalinglanguagemodeling, touvron2023llamaopenefficientfoundation, qwen2025qwen25technicalreport}. Post-training ~\citep{openai2024openaio1card, deepseekai2025deepseekr1incentivizingreasoningcapability} and test-time scaling strategies~\citep{muennighoff2025s1simpletesttimescaling} enable sophisticated reasoning behaviors in LLMs and extend the length of Chain-of-Thoughts (CoTs)~\citep{wei2022chainofthought}, thereby achieving strong results on challenging benchmarks~\citep{di_zhang_2025, penedo2025codeforces}. A key component of these advances is the Process Reward Models (PRMs)~\citep{lightman2023let, li2023makinglargelanguagemodels}, which provide fine-grained, step-wise supervision of the reasoning process and reliable selection of high-quality reasoning trajectories. These developments are proven highly effective for improving the performance of LLMs in complex tasks~\citep{ma2023letsrewardstepstep, wang2024qimprovingmultistepreasoning}.

Given the success with LLMs, a natural extension is to apply PRMs to multimodal large language models (MLLMs)~\citep{wu2023multimodallargelanguagemodels,li2025survey}  to enhance their reasoning abilities. Early studies of multimodal PRMs demonstrate promise results, yet substantial challenges persist. Distinct from text-only inputs of LLMs, MLLMs must combine diverse visual and language signals: a high-dimensional, continuous image space coupled with discrete language tokens. This fusion dramatically broadens the input manifold and leads to more severe \emph{distribution shifts}~\citep{song2025bridgegapmodalitiessurvey} from training to testing distributions. Consequently, directly utilizing PRM training strategies from the text domain~\citep{wang2024multistepproblemsolvingverifier,luo2024improvemathematicalreasoninglanguage} underperforms, mainly due to the decreased generalizability~\citep{dong2024progressivemultimodalreasoningactive} caused by the insufficient coverage of the multimodal input space. 

A straightforward solution to this problem is to combine multiple datasets that emphasize different multimodal reasoning skills, thereby enlarging the sampling space. However, \emph{quality imbalance} among existing multimodal reasoning datasets is more severe than in text-only settings: many contain noisy inputs such as unnecessary modalities~\citep{yu2024mmvetevaluatinglargemultimodal} or questions of negligible difficulty~\citep{lu2024mathvista}, as illustrated in Fig.~\ref{fig:intro}. Since these easy datasets contribute little to effective sampling, paying much attention to them can substantially degrade PRM performance. Therefore, an effective data selection strategy that filters out unreliable datasets and instances is crucial to training a high-quality multimodal PRM.

To overcome these challenges, we propose DreamPRM, a domain-reweighted training framework for multimodal PRMs. Inspired by domain-reweighting techniques~\citep{shu2019metaweightnetlearningexplicitmapping,fan2024dogedomainreweightinggeneralization,sow2025dynamiclossbasedsamplereweighting}, DreamPRM dynamically learns appropriate weights for each multimodal reasoning dataset, allowing them to contribute unequally during training. Datasets that contain many noisy samples tend to receive lower domain weights, reducing their influence on PRM parameter updates. Conversely, high-quality datasets are assigned higher weights and thus play a more important role in optimization. This domain-reweighting strategy alleviates the issue of dataset quality imbalances. DreamPRM adopts a bi-level optimization (BLO) framework~\citep{Finn2017ModelAgnosticMF,liu2018darts} to jointly learn the domain weights and PRM parameters. At the lower level, the PRM parameters are optimized with Monte Carlo signals on multiple training domains under different domain weights. At the upper level, the optimized PRM is evaluated on a separate meta domain to compute a novel aggregation function loss, which is used to optimized the domain weights. Extensive experiments on a wide range of multimodal reasoning benchmarks verify the effectiveness of DreamPRM.

Our contributions are summarized as follows:
\begin{itemize}
\item We propose DreamPRM, a \emph{domain-reweighted} multimodal process reward model training framework that dynamically adjusts the importance of different training domains. We formulate the training process of DreamPRM as a \emph{bi-level optimization} (BLO) problem, where the lower level optimizes the PRM via domain-reweighted fine-tuning, and the upper level optimizes domain weights with an aggregation function loss. Our method helps address dataset quality imbalance issue in multimodal reasoning, and improves the generalization ability of PRM.
\item We conduct extensive experiments using DreamPRM on a wide range of multimodal reasoning benchmarks. Results indicate that DreamPRM consistently surpasses PRM baselines with other data selection strategies, confirming the effectiveness of its bi-level optimization based domain-reweighting strategy. Notably, DreamPRM achieves a top-1 accuracy of 85.2\% on the \textsc{MathVista} leaderboard using the o4-mini model, demonstrating its strong generalization in complex multimodal reasoning tasks. Carefully designed evaluations further demonstrate that DreamPRM possesses both scaling capability and generalization ability to stronger models.
\end{itemize}

\section{Related Works}\label{app:related_work}
\paragraph{Multimodal reasoning} Recent studies have demonstrated that incorporating Chain-of-Thought (CoT) reasoning~\citep{wei2022chain, NEURIPS2022_8bb0d291, zhang2023automatic} into LLMs encourages a step-by-step approach, thereby significantly enhancing question-answering performance. However, it has been reported that CoT prompting can't be easily extended to MLLMs, mainly due to hallucinated outputs during the reasoning process~\citep{wang2024mpo, zheng2024thinkinglookingimprovingmultimodal, jiang2025mmecotbenchmarkingchainofthoughtlarge}. Therefore, some post-training methods have been proposed for enhancing reasoning capability of MLLMs. InternVL-MPO~\citep{wang2024mpo} proposes a mixed preference optimization that jointly optimizes preference ranking, response quality, and response generation loss to improve the reasoning abilities. Llava-CoT~\citep{xu2024llavacot} creates a structured thinking fine-tuning dataset to make MLLM to perform systematic step-by-step reasoning. Some efforts have also been made for inference time scaling. RLAIF-V~\citep{yu2024rlaifv} proposes a novel self-feedback guidance for inference-time scaling and devises a simple length-normalization strategy tackling the bias towards shorter responses. AR-MCTS~\citep{dong2024progressivemultimodalreasoningactive} combines Monte-Carlo Tree Search (MCTS) and Retrival Augmented Generation (RAG) to guide MLLM search step by step and explore the answer space.

\paragraph{Process reward model} Process Reward Model (PRM)~\citep{lightman2023let, li2023makinglargelanguagemodels, ma2023letsrewardstepstep, wang2024qimprovingmultistepreasoning} provides a more finer-grained verification than Outcome Reward Model (ORM)~\citep{cobbe2021trainingverifierssolvemath, shao2024deepseekmathpushinglimitsmathematical}, scoring each step of the reasoning trajectory. However, a central challenge in designing PRMs is obtaining process supervision signals, which require supervised labels for each reasoning step. Current approaches typically depend on costly, labor-intensive human annotation~\citep{lightman2023let}, highlighting the need for automated methods to improve scalability and efficiency. Math-Shepherd~\citep{wang-etal-2024-math} proposes a method utilizing Monte-Carlo estimation to provide hard labels and soft labels for automatic process supervision. OmegaPRM~\citep{luo2024improvemathematicalreasoninglanguage} proposes a Monte Carlo Tree Search (MCTS) for finer-grained exploration for automatical labeling. MiPS~\citep{wang2024multistepproblemsolvingverifier} further explores the Monte Carlo estimation method and studies the aggregation of PRM signals.

\paragraph{Domain-reweighting} Domain reweighting methodologies are developed to modulate the influence of individual data domains, thereby enabling models to achieve robust generalization. Recently, domain reweighting has emerged as a key component in large language model pre-training, where corpora are drawn from heterogeneous sources. DoReMi~\citep{xie2023doremi} trains a lightweight proxy model with group distributionally robust optimization to assign domain weights that maximize excess loss relative to a reference model. DOGE~\citep{DOGE} proposes a first-order bi-level optimization framework, using gradient alignment between source and target domains to update mixture weights online during training. Complementary to these optimization-based approaches, Data Mixing Laws~\citep{ye2025data} derives scaling laws that could predict performance under different domain mixtures, enabling low-cost searches for near-optimal weights without proxy models. In this paper, we extend these ideas to process supervision and introduce a novel bi-level domain-reweighting framework.
\section{Problem Setting and Preliminaries}
\label{sec:prob_set_prelim}
\paragraph{Notations.} Let $\mathcal{I}$, $\mathcal{T}$, and $\mathcal{Y}$ denote the multimodal input space (images), textual instruction space, and response space, respectively. A multimodal large language model (MLLM) is formalized as a parametric mapping $M_\theta: \mathcal{T} \times \mathcal{I} \to \Delta(\mathcal{Y})$, where $\hat{y} \sim M_\theta(\cdot|x)$ represents the stochastic generation of responses conditioned on input pair $x=(t, I)$ including visual input $I \in \mathcal{I}$ and textual instruction $t \in \mathcal{T}$, with $\Delta(\mathcal{Y})$ denoting the probability simplex over the response space. We use $y\in \mathcal{Y}$ to denote the ground truth label from a dataset.

The process reward model (PRM) constitutes a sequence classification function $\mathcal{V}_\phi: \mathcal{T} \times \mathcal{I} \times \mathcal{Y} \to [0, 1]$, parameterized by $\phi$, which quantifies the epistemic value of partial reasoning state $\hat{y}_i$ through scalar reward $p_i = \mathcal{V}_\phi(x,\hat{y}_i)$, modeling incremental utility toward solving instruction $t$ under visual grounding $I$. Specifically, $\hat{y}_i$ represents the first \(i\) steps of a complete reasoning trajectory $\hat{y}$.



\paragraph{PRM training with Monte Carlo signals.} Due to the lack of ground truth epistemic value for each partial reasoning state $\hat{y}_i$, training of PRM requires automatic generation of approximated supervision signals. An effective approach to obtain these signals is to use the Monte Carlo method~\citep{wang2024multistepproblemsolvingverifier, wang-etal-2024-Math-Shepherd}. We first feed the input question-image pair \(x=(t, I)\) and the prefix solution \(\hat{y}_i\) into the MLLM, and let it complete the remaining steps until reaching the final answer. We randomly sample multiple completions, compare their final answers to the gold answer $y$, and thereby obtain multiple correctness labels. PRM is trained as a sequence classification task to predict these correctness labels. The ratio of correct completions at the \(i\)-th step estimates the ``correctness level'' up to step \(i\), which is used as the approximated supervision signals $p_i$ to train the PRM. Formally,

\begin{equation}
p_i = \texttt{MonteCarlo}(x, \hat{y}_i, y)
 = \frac{\texttt{num(correct completions from } \hat{y}_i)}{\texttt{num(total completions from } \hat{y}_i)}
\label{eq:montecarlo}
\end{equation}

\paragraph{PRM-based inference with aggregation function.}
\label{sec:prm-infer} 
\label{sec:prob_setting}
\begin{figure}[t]
  \centering
  \includegraphics[width=\textwidth]{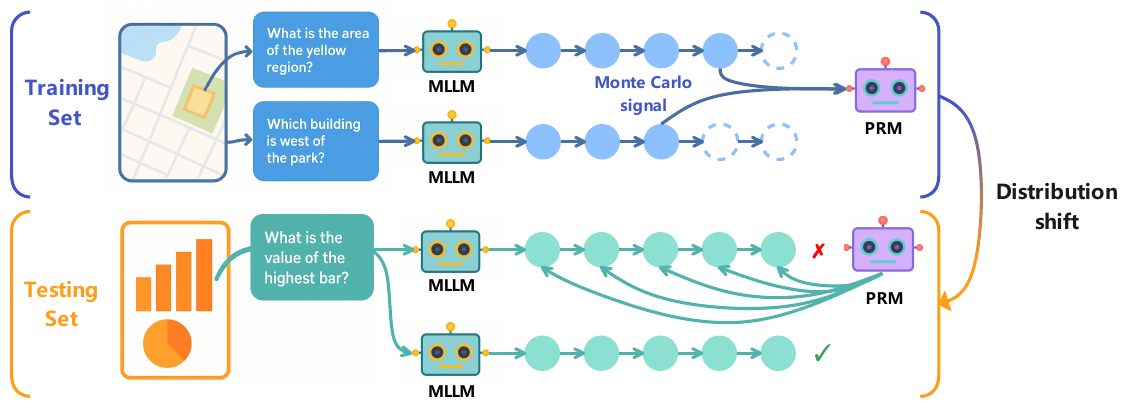} 
  \caption{\textbf{General flow of training PRM and using PRM for inference. Training phase}: Train PRM with Monte Carlo signals from intermediate steps of Chain-of-Thoughts (CoTs). \textbf{Inference phase}: Use the trained PRM to verify CoTs step by step and select the best CoT. Conventional training of PRM has poor generalization capability due to \emph{distribution shift} between training set and testing set.}
  \label{fig:prm}
\end{figure}

After training a PRM, a typical way of conducting PRM-based MLLM inference is to use aggregation function~\citep{wang2024multistepproblemsolvingverifier}. Specifically, for each candidate solution $\hat{y}$ from the MLLM, PRM will generate a list of predicted probabilities ${p}=\{{p_1}, {p_2}, . . . , {p_n}\}$ accordingly, one for each step $\hat{y}_i$ in the solution. The list of predicted probabilities are then aggregated using the following function: 

\begin{equation}
\label{eq:agg}
\mathcal A({p})= \sum_{i=1}^{n}\log\frac{{p_i}}{1-{p_i}}.
\end{equation}

The aggregated value corresponds to the score of a specific prediction $\hat{y}$, and the final PRM-based solution is the one with the highest aggregated score.

\paragraph{Bi-level optimization.} Bi-level optimization (BLO) has been widely used in meta-learning~\citep{Finn2017ModelAgnosticMF}, neural architecture search~\citep{liu2018darts}, and data reweighting~\citep{Meta-Weight-Net}. A BLO problem is usually formulated as:

\begin{align}
    &\min_\alpha \mathcal U(\alpha, \phi^*(\alpha)) \\
    s.t. & \phi^*(\alpha) = \underset{\mathbf{\phi}}{\arg \min } \mathcal L(\phi, \alpha)
\end{align}

where $\mathcal U$ is the upper-level optimization problem (OP) with parameter $\alpha$, and $\mathcal L$ is the lower-level OP with parameter $\phi$. The lower-level OP is nested within the upper-level one, and the two OPs are mutually dependent.

\section{The Proposed Domain-reweighting Method}
\label{sec:method}

\begin{figure}[t]
  \centering
  \includegraphics[width=\textwidth]{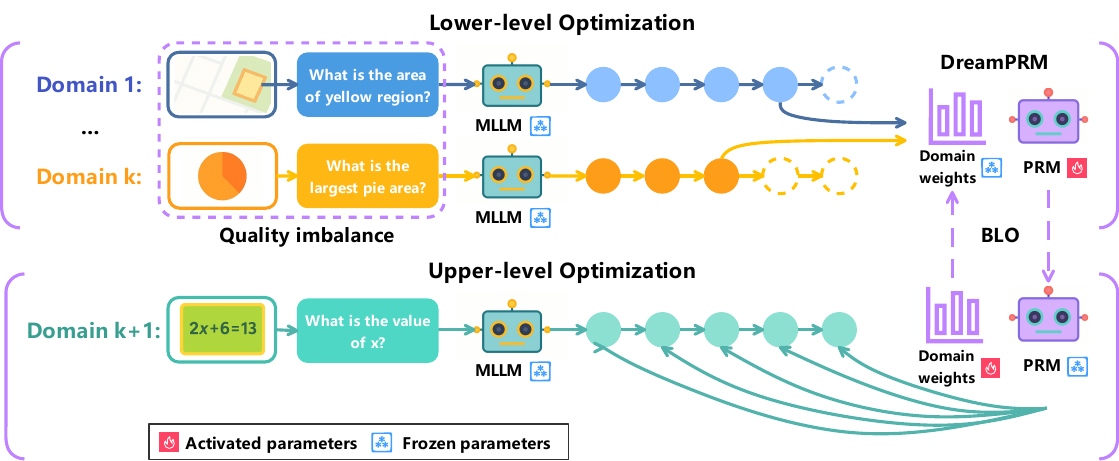}
  \caption{\textbf{The proposed bi-level optimization based domain-reweighting method.} \textbf{Lower-level optimization:} In this stage, PRM's parameters are updated on multiple datasets with domain weights, allowing the PRM to prioritize domains with better quality. \textbf{Upper-level optimization:} In this stage,  the PRM is evaluated on a separate meta dataset to compute an aggregation function loss and optimize the domain weights. DreamPRM helps address dataset {quality imbalance} problems and leads to stronger and more generalizable reasoning performance.}
  \label{fig:method}
\end{figure}

\paragraph{Overview.} 

Training process reward models (PRMs) for MLLMs is challenging for two reasons: (1) dataset (domain) quality imbalance, and (2) discrepancy between training and inference procedures. To address these two challenges, we propose DreamPRM, which automatically searches for domain importance using a novel aggregation function loss that better simulates the inference process of PRM. Under a bi-level optimization framework, it optimizes PRM parameters with Monte Carlo signals at the lower level, and optimizes trainable domain importance weights with aggregation function loss at the upper level. An overview of DreamPRM method is shown in Fig. \ref{fig:method}.

\paragraph{Datasets.} We begin with $K{+}1$ datasets, each from a distinct domain (e.g., science, geometry).  
The first $K$ datasets form the training pool
$\mathcal{D}_{\mathrm{tr}}=\{\mathcal{D}_1,\dots,\mathcal{D}_K\}$,  
while the remaining dataset, $\mathcal{D}_{\mathrm{meta}}=\mathcal{D}_{K+1}$, is a meta (validation) dataset with better quality.

\paragraph{Lower-level optimization: domain-reweighted training of PRM.} In lower-level optimization, we aim to update the weights $\phi$ of PRM with domain-reweighted training. We first define the typical PRM training loss $\mathcal L_{tr}$ on a single domain $\mathcal D_k$, given PRM parameters $\phi$, as follows:

\begin{align}
    \mathcal L_{tr} (\mathcal D_k, \phi) = \sum_{(x, y)\in\mathcal D_k} \sum_{i=1}^n \mathcal{L}_{MSE}(\mathcal{V}_\phi(x, \hat{y}_i), p_i)
\end{align}

where $\hat{y}_i$ is the prefix of MLLM generated text $\hat{y}=M_\theta(x)$ given input pair $x=(t, I)$, and $p_i$ is the process supervision signal value obtained by Monte Carlo estimation given input pair $x$, prefix $\hat{y}_i$ and ground truth label $y$, as previously defined in Equation \ref{eq:montecarlo}. The PRM is optimized by minimizing the mean squared error (MSE) between supervision signal and PRM predicted score $\mathcal{V}_\phi(x, \hat{y}_i)$. With the PRM training loss on a single domain $\mathcal D_k$ above, we next define the domain-reweighted training objective of PRM on multiple training domains $\mathcal{D}=\{\mathcal{D}_k\}_{k=1}^K$. The overall objective is a weighted sum of the single-domain PRM training losses, allowing the contribution of each domain to be adjusted during the learning process:

\begin{align}
    \mathcal{L}_{tr}(\mathcal D_{tr}, \phi, \alpha)=\sum_{k=1}^{K} \alpha_k\mathcal{L}_{tr}(\mathcal D_k, \phi) 
\end{align}

Here, $\alpha = \{\alpha_k\}_{k=1}^K$ represents the trainable domain weight parameters, indicating the importance of each domain. By optimizing this objective, we obtain the optimal value of PRM parameters $\phi^*$: 

\begin{align}
    \phi^*(\alpha) = & \underset{\mathbf{\phi}}{\arg \min } \mathcal{L}_{tr}(\mathcal D_{tr}, \phi, \alpha) 
\end{align}

It is worth mentioning that only $\phi$ is optimized at this level, while $\alpha$ remains fixed.





\paragraph{Upper-level optimization: learning domain reweighting parameters.} In upper-level optimization, we optimize the domain reweighting parameter $\alpha$ on meta dataset $\mathcal D_{meta}$ given optimal PRM weights $\phi^*(\alpha)$ obtained from the lower level. To make the meta learning target more closely reflect the actual PRM-based inference process, we propose a novel meta loss function $\mathcal L_{meta}$, different from the training loss $\mathcal L_{tr}$. Specifically, we first obtain an aggregated score $\mathcal A({p})$ for each generated solution $\hat{y}$ from the MLLM given input pair $x=(t, I)$, following process in Section \ref{sec:prm-infer}. We then create a ground truth signal $r(\hat{y}, y)$ by assigning it a value of 1 if the generated $\hat{y}$ contains ground truth $y$, and 0 otherwise. The meta loss is defined as the mean squared error between aggregated score and ground truth signal:

\begin{align}
    \mathcal L_{meta} (\mathcal D_{meta}, \phi^*(\alpha)) = \sum_{(x, y)\in\mathcal D_{meta}} \mathcal{L}_{MSE}(\sigma(\mathcal A(\mathcal{V}_{\phi^*(\alpha)}(x, \hat{y}))), r(\hat{y}, y))
\end{align}

where $\mathcal A$ represents the aggregation function as previously defined in Equation \ref{eq:agg}, and $\sigma$ denotes the sigmoid function to map the aggregated score to a probability. Accordingly, the optimization problem at the upper level is formulated as follows:

\begin{align}
\label{eq:upper}
    \underset{\alpha}{\min }\mathcal L_{meta}(\mathcal D_{meta}, \phi^*(\alpha)) 
\end{align}

To solve this optimization problem, we propose an efficient gradient-based algorithm, which is detailed in Appendix \ref{app:optim-algo}.


\section{Experimental Results}
\label{sec:exp}

\subsection{Experimental settings} 

\label{sec:expr-setting}
\paragraph{Multistage reasoning.}
To elicit consistent steady reasoning responses from current MLLMs, we draw on the Llava-CoT approach \citep{xu2025llavacotletvisionlanguage}, which fosters structured thinking prior to answer generation. Specifically, we prompt MLLMs to follow five reasoning steps:  \texttt{(1) Restate the question. (2) Gather evidence from the image. (3) Identify any background knowledge needed. (4) Reason with the current evidence. (5) Summarize and conclude with all the information.} We also explore zero-shot prompting settings in conjunction with structural reasoning, which can be found in Appendix~\ref{app:st}. We use 8 different chain-of-thought reasoning trajectories for all test-time scaling methods, unless otherwise stated.




\begin{table}[t]
\centering
\caption{\textbf{Comparative evaluation of DreamPRM and baselines on multimodal reasoning benchmarks.}  
\textbf{Bold numbers} indicate the best performance, while \underline{underlined numbers} indicate the second best.  
The table reports accuracy (\%) on five datasets: \textsc{WeMath}, \textsc{MathVista}, \textsc{MathVision}, \textsc{MMVet}, and \textsc{MMStar}.}

\renewcommand{\arraystretch}{1.2} 
\resizebox{\textwidth}{!}{
\begin{tabular}{lccccc}
\toprule
& \multicolumn{3}{c}{\textbf{Math Reasoning}} & \multicolumn{2}{c}{\textbf{General Reasoning}} \\
\cmidrule(lr){2-4} \cmidrule(lr){5-6}
\textbf{} & \textsc{WeMath} & \textsc{MathVista} & \textsc{MathVision} & \textsc{MMVet} & \textsc{MMStar}\\
&\textit{(loose)} &\textit{(testmini)} &\textit{(test)} &\textit{(v1)} &\textit{(test)}\\
\midrule
\multicolumn{6}{c}{\textit{Zero-shot Methods}} \\
\midrule
Gemini-1.5-Pro~\citep{reid2024gemini}       & 46.0 & 63.9 & 19.2 & \underline{64.0} & 59.1    \\
GPT-4v~\citep{openai2024gpt4technicalreport}               & 51.4 & 49.9 & \underline{21.7} & \bf67.7 & \underline{62.0}    \\
LLaVA-OneVision-7B~\citep{li2024llavaonevisioneasyvisualtask}         & 44.8    & 63.2   & 18.4   & 57.5   & 61.7    \\
Qwen2-VL-7B~\citep{Qwen2VL}               & 42.9    & 58.2   & 16.3   & 62.0   & 60.7    \\ 
InternVL-2.5-8B-MPO~\citep{wang2024mpo}             & 51.7    & 65.4   & 20.4   & 55.9   & 58.9    \\
\midrule
\multicolumn{6}{c}{\textit{Test-time Scaling Methods (InternVL-2.5-8B-MPO based)}} \\
\midrule
Self-consistency~\citep{wang2023selfconsistencyimproveschainthought}        & 56.4    & 67.1   & 20.7   & 57.4   & 59.6    \\
Self-correction~\citep{he2024selfcorrectionrefinementlearningframework}         & 54.0    & 63.8   & 21.6   & 54.9   & 59.7    \\
ORM~\citep{shao2024deepseekmathpushinglimitsmathematical}  & 56.9  & 65.3  & 20.5   & 55.9   & 60.1    \\
\midrule
Vanilla PRM~\citep{lightman2023let}        & 54.2    & 67.2   & 20.6   & 58.9   & 60.8 \\
CaR-PRM~\citep{ge2024clusteringrankingdiversitypreservedinstruction}      & 54.7    & \underline{67.5}   & 21.0   & 60.6   & 61.1\\
s1-PRM~\citep{muennighoff2025s1simpletesttimescaling} & \underline{57.1}    & 65.8   & 20.2   & 60.1   & 60.4 \\

\textbf {DreamPRM (ours)}     &\bf 57.4    &\bf  68.9   & \bf22.1   & 61.4   & \bf62.3    \\

\bottomrule
\end{tabular}}
\label{tab:main}
\end{table}

\paragraph{Base models.} For inference, we use InternVL-2.5-8B-MPO~\citep{wang2024mpo} as the base MLLM, which has undergone post-training to enhance its reasoning abilities and is well-suited for our experiment. For fine-tuning PRM, we adopt Qwen2-VL-2B-Instruct~\citep{Qwen2VL}. Qwen2-VL is a state-of-the-art multimodal model pretrained for general vision-language understanding tasks. This pretrained model serves as the initialization for our fine-tuning process.

\paragraph{Training hyperparameters.}\label{app:training}  
 In the lower-level optimization,  we perform 5 inner gradient steps per outer update ({unroll steps} = 5) using the AdamW~\citep{loshchilov2019decoupledweightdecayregularization} optimizer with learning rate set to $5\times10^{-7}$.  
In the upper-level optimization, we use the {AdamW} optimizer ($\mathrm{lr}=0.01$, weight decay $=10^{-3}$) and a StepLR scheduler (step size = 5000, $\gamma=0.5$). In total, DreamPRM is fine-tuned for 10000 iterations. Our method is implemented with {Betty}~\citep{choe2023betty}, and the fine-tuning process takes approximately 10 hours on one NVIDIA A100 GPUs. 

\paragraph{Baselines.} We use three major categories of baselines: \textbf{(1)} State-of-the-art models on public leaderboards, including Gemini-1.5-Pro~\citep{reid2024gemini}, GPT-4V~\cite{openai2024gpt4technicalreport}, LLaVA-OneVision-7B~\citep{li2024llavaonevisioneasyvisualtask}, Qwen2-VL-7B~\citep{Qwen2VL}. We also carefully reproduce the results of InternVL-2.5-8B-MPO with structural thinking. \textbf{(2)} Test-time scaling methods (excluding PRM) based on the InternVL-2.5-8B-MPO model, including: (i) Self-consistency~\citep{wang2023selfconsistencyimproveschainthought}, which selects the most consistent reasoning chain via majority voting over multiple responses; (ii) Self-correction~\citep{he2024selfcorrectionrefinementlearningframework}, which prompts the model to critically reflect on and revise its initial answers; and (iii) Outcome Reward Model (ORM)~\citep{shao2024deepseekmathpushinglimitsmathematical}, which evaluates and scores the final response to select the most promising one. \textbf{(3)} PRM-based methods, including: (i) Vanilla PRM trained without any data selection, as commonly used in LLM settings~\citep{lightman2023let}; (ii) s1-PRM, which selects high-quality reasoning responses based on three criteria - difficulty, quality, and diversity - following the s1 strategy~\citep{muennighoff2025s1simpletesttimescaling}; and (iii) CaR-PRM, which filters high-quality visual questions using clustering and ranking techniques, as proposed in CaR~\citep{ge2024clusteringrankingdiversitypreservedinstruction}.

\paragraph{Datasets and benchmarks.} We use 15 multimodal datasets for lower-level optimization ($\mathcal{D}_{tr}$), covering four domains: science, chart, geometry, and commonsense, as listed in Appendix Table~\ref{tab:vqa_datasets}. For upper-level optimization ($\mathcal{D}_{meta}$), we adopt the \textsc{MMMU}~\citep{yue2024mmmumassivemultidisciplinemultimodal} dataset. Evaluation is conducted on five multimodal reasoning benchmarks: \textsc{WeMath}~\citep{qiao2024wemathdoeslargemultimodal}, \textsc{MathVista}~\citep{lu2024mathvista}, \textsc{MathVision}~\citep{wang2024measuringmultimodalmathematicalreasoning}, \textsc{MMVet}~\citep{yu2024mmvetevaluatinglargemultimodal}, and \textsc{MMStar}~\citep{chen2024rightwayevaluatinglarge}. Details are provided in Appendix~\ref{sec:append_1}.

\subsection{Benchmark evaluation of DreamPRM}
Tab.~\ref{tab:main} presents the primary experimental results. We observe that: \textbf{(1) DreamPRM outperforms other PRM-based methods}, highlighting the effectiveness of our domain reweighting strategy. Compared to the vanilla PRM trained without any data selection, DreamPRM achieves a consistent performance gain of 2\%-3\% across all five datasets, suggesting that effective data selection is crucial for training high-quality multimodal PRMs. Moreover, DreamPRM also outperforms s1-PRM and CaR-PRM, which rely on manually designed heuristic rules for data selection. These results indicate that selecting suitable reasoning datasets for PRM training is a complex task, and handcrafted rules are often suboptimal. In contrast, our automatic domain-reweighting approach enables the model to adaptively optimize its learning process, illustrating how data-driven optimization offers a scalable solution to dataset selection challenges. \textbf{(2) DreamPRM outperforms SOTA MLLMs with much fewer parameters}, highlighting the effectiveness of DreamPRM. For example, DreamPRM significantly surpasses two trillion-scale closed-source LLMs (GPT-4v and Gemini-1.5-Pro) on 4 out of 5 datasets. In addition, it consistently improves the performance of the base model, InternVL-2.5-8B-MPO, achieving an average gain of 4\% on the five datasets. These results confirm that DreamPRM effectively yields a high-quality PRM, which is capable of enhancing multimodal reasoning across a wide range of benchmarks. \textbf{(3) DreamPRM outperforms other test-time scaling methods}, primarily because it enables the training of a high-quality PRM that conducts fine-grained, step-level evaluation. While most test-time scaling methods yield moderate improvements, DreamPRM leads to the most substantial gains, suggesting that the quality of the reward model is critical for effective test-time scaling. We further provide case studies in Appendix~\ref{app:additional}, which intuitively illustrate how DreamPRM assigns higher scores to coherent and high-quality reasoning trajectories.

\begin{figure}[t]
  \centering
  \includegraphics[width=\textwidth]{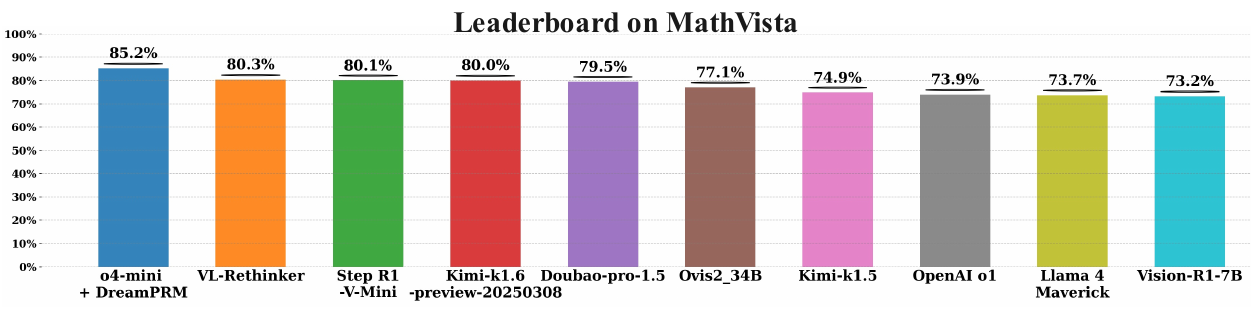}
  \caption{
\textbf{Leaderboard on MathVista (as of October 15, 2025).}
The first column (``o4-mini + DreamPRM'') reports our own evaluation,
while the remaining results are taken from the official MathVista leaderboard.
The compared models include VL-Rethinker~\citep{Wang2025VLRethinker},
Step R1-V-Mini~\citep{StepFun2025StepR1VMini},
Kimi-k1.6-preview~\citep{Kimi2025k16preview},
Kimi-k1.5~\citep{Kimi2025k15},
Doubao-pro-1.5~\citep{DoubaoProduct},
Ovis2-34B~\citep{Ovis2_34B_2025},
OpenAI~o1~\citep{openai2024openaio1card},
Llama~4~Maverick~\citep{Llama4Maverick2025Blog,Llama4MaverickHF},
and Vision-R1-7B~\citep{Huang2025VisionR1}.
}
  \label{fig:leaderboard_mathvista}
\end{figure}

\subsection{Leaderboard performance of DreamPRM}

As shown in Fig.~\ref{fig:leaderboard_mathvista}, DreamPRM achieves the \textbf{top-1 accuracy of 85.2\%} on the \textsc{MathVista} leaderboard (as of October 15, 2025). 
The result (o4-mini + DreamPRM) has been officially verified through the MathVista evaluation. 
Compared with a series of strong multimodal reasoning baselines, including 
VL-Rethinker~\citep{Wang2025VLRethinker}, 
Step R1-V-Mini~\citep{StepFun2025StepR1VMini}, 
Kimi-k1.6-preview~\citep{Kimi2025k16preview}, 
Doubao-pro-1.5~\citep{DoubaoProduct}, 
Ovis2-34B~\citep{Ovis2_34B_2025}, 
OpenAI~o1~\citep{openai2024openaio1card}, 
Llama~4~Maverick~\citep{Llama4Maverick2025Blog,Llama4MaverickHF}, 
and Vision-R1-7B~\citep{Huang2025VisionR1}, 
DreamPRM demonstrates clearly superior multimodal reasoning capability.

Table~\ref{tab:prm_comparison} in Appendix provides a detailed comparison among various Process Reward Model (PRM) variants built on the same o4-mini backbone. 
DreamPRM surpasses all counterparts, improving the base o4-mini model from 80.6\% (pass@1) and 81.7\% (self-consistency@8) to 85.2\%. 
This consistent gain verifies the effectiveness of DreamPRM in enhancing reasoning accuracy through process-level supervision and reliable consensus across multiple chains of thought.

\begin{figure}[t]
  \centering
  \includegraphics[width=1.\textwidth]{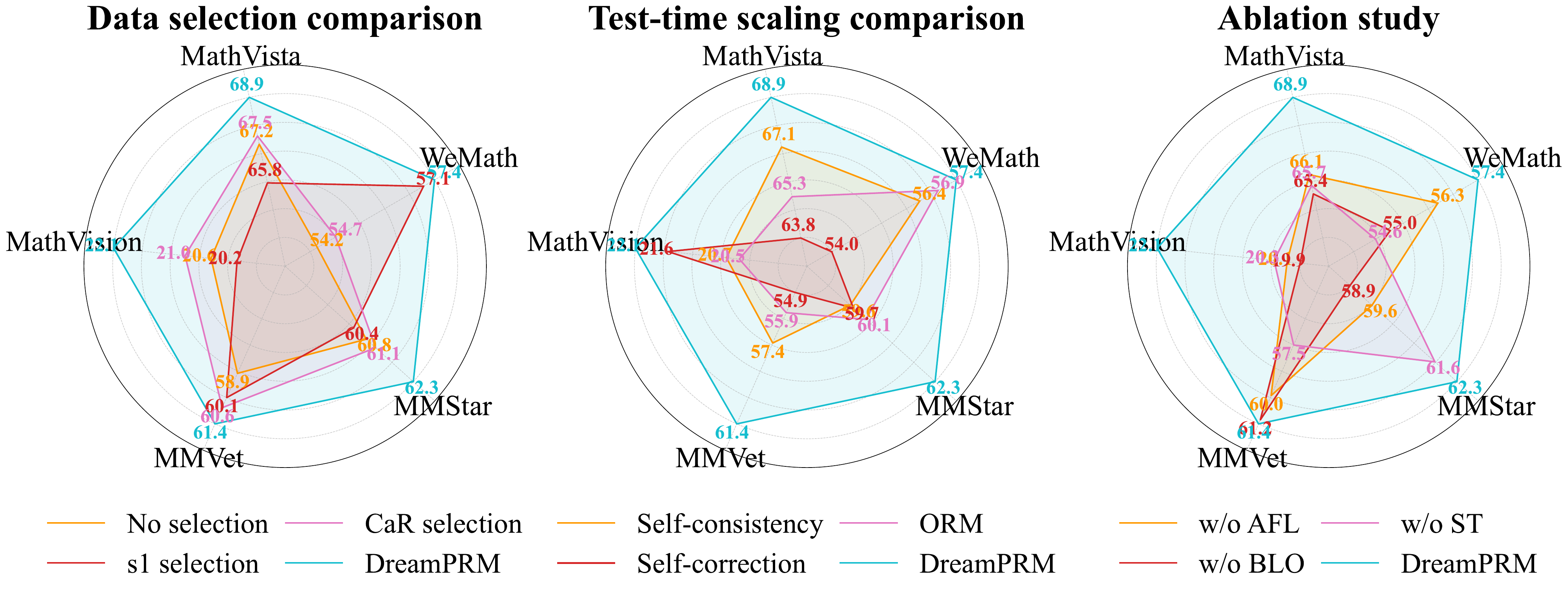} 
  \caption{\textbf{Comparative evaluation of DreamPRM on multimodal reasoning benchmarks.}  
Radar charts report accuracy (\%) on five datasets (\textsc{WeMath}, \textsc{MathVista}, \textsc{MathVision}, \textsc{MMVet}, and \textsc{MMStar}).  
\textbf{(a)}~Impact of different data selection strategies.  
\textbf{(b)}~Comparison with existing test-time scaling methods.  
\textbf{(c)}~Ablation study of three key components, i.e. w/o aggregation function loss (AFL), w/o bi-level optimization (BLO), and w/o structural thinking (ST).}
  \label{fig:results}
\end{figure}

\begin{figure}[t]
    \centering
    \begin{subfigure}[b]{0.33\textwidth}
        \centering
        \includegraphics[width=\linewidth]{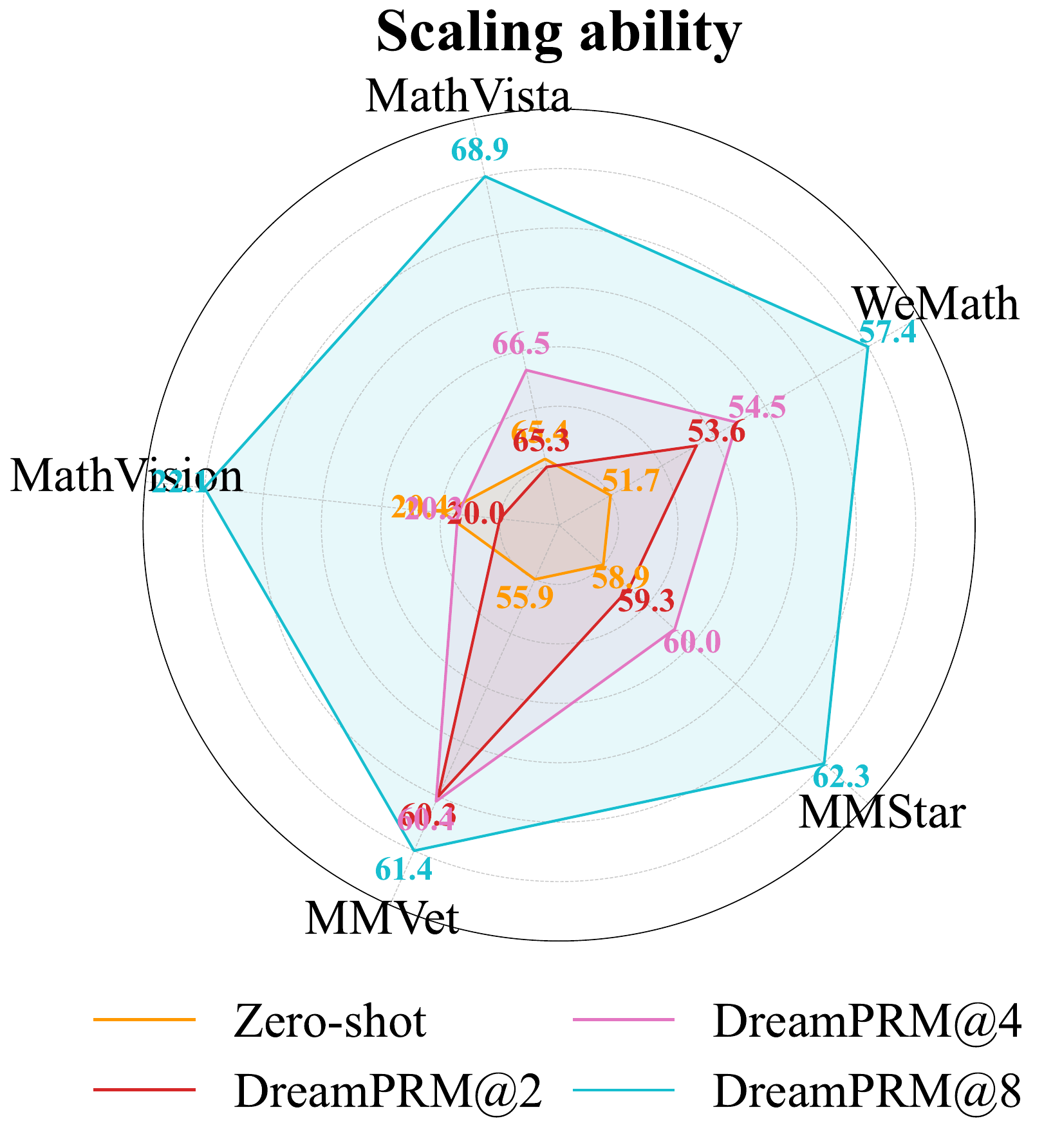}
        \label{fig:img1}
    \end{subfigure}
    \hfill
    \begin{subfigure}[b]{0.57\textwidth}
        \centering
        \includegraphics[width=\linewidth]{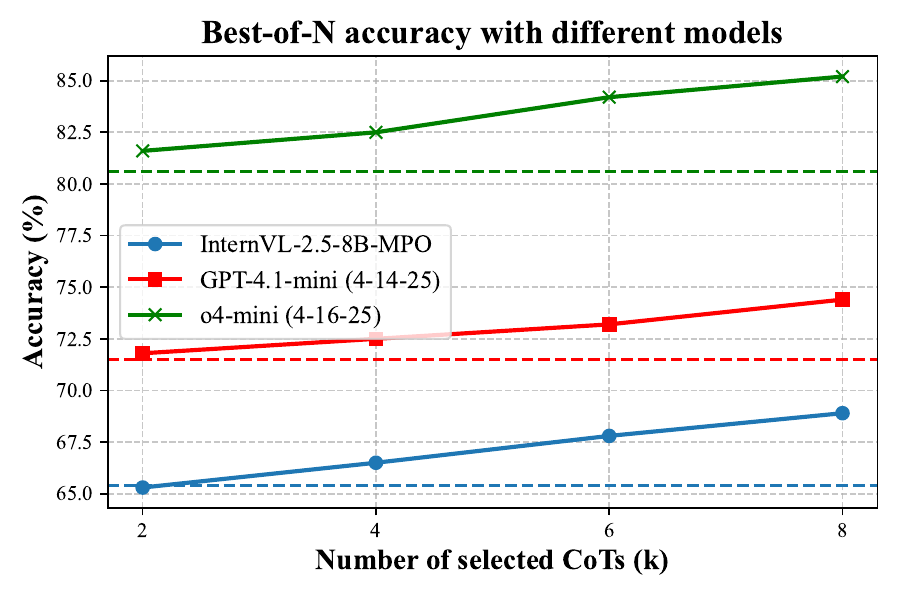}
        \label{fig:img2}
    \end{subfigure}
    \caption{\textbf{Scaling ability and cross-model generalization.}
\textbf{(a)}~Radar chart of five multimodal reasoning benchmarks shows that DreamPRM delivers monotonic accuracy gains as the number of selected chains-of-thought increases (\mbox{@2}, \mbox{@4}, \mbox{@8}) over the pass@1 baseline.  
\textbf{(b)}~Best-of-\textit{N} accuracy curves for InternVL-2.5-8B-MPO (blue), GPT-4.1-mini (red) and o4-mini (green) on \textsc{MathVista} confirm that the same DreamPRM-ranked CoTs generalize across models, consistently outperforming pass@1 performance (dashed lines) as $k$ grows.}
    \label{fig:scaling}
\end{figure}

\subsection{Scaling and generalization analysis of DreamPRM}

\textbf{DreamPRM scales reliably with more CoT candidates.} As shown in the left panel of Fig.~\ref{fig:scaling}, the accuracy of DreamPRM consistently improves on all five benchmarks as the number of CoTs increases from $k{=}2$ to $k{=}8$, expanding the radar plot outward. Intuitively, a larger set of candidates increases the likelihood of including high-quality reasoning trajectories, but it also makes identifying the best ones more challenging. The consistent performance gains indicate that DreamPRM effectively verifies and ranks CoTs, demonstrating its robustness in selecting high-quality reasoning trajectories under more complex candidate pools.

\textbf{DreamPRM transfers seamlessly to stronger base MLLMs.}  The right panel of Fig.\ref{fig:scaling} shows the \textsc{MathVista} accuracy when applying DreamPRM to recent MLLMs, GPT-4.1-mini \textit{(2025-04-14)}~\citep{openai2024gpt4technicalreport} and o4-mini \textit{(2025-04-16)}~\citep{openai2024openaio1card}. For o4-mini model, the pass@1 score of 80.6\% steadily increases to 85.2\% at $k{=}8$, surpassing the previous state-of-the-art performance. This best-of-$N$ trend, previously observed with InternVL, also holds for GPT-4.1-mini and o4-mini, demonstrating the generalization ability of DreamPRM. Full results of these experiments are provided in Tab.~\ref{tab:bestofN}.

\subsection{Ablation study} 

In this section, we investigate the importance of three components in DreamPRM: \textbf{(1)} bi-level optimization, \textbf{(2)} aggregation function loss in upper-level, and \textbf{(3)} structural thinking prompt (detailed in Section \ref{sec:expr-setting}). As shown in the rightmost panel of Fig. \ref{fig:results}, the complete DreamPRM achieves the best results compared to three ablation baselines across all five benchmarks. Eliminating bi-level optimization causes large performance drop (e.g., -3.5\% on \textsc{MathVista} and -3.4\% on \textsc{MMStar}). Removing aggregation function loss leads to a consistent 1\%-2\% decline (e.g., 57.4\% $\rightarrow$ 56.3\% on \textsc{WeMath}). Excluding structural thinking also degrades performance (e.g., -1.8\% on \textsc{MathVision}). These results indicate that all three components are critical for DreamPRM to achieve the best performance. More detailed results are shown in Appendix Tab.~\ref{tab:ablation}.

\subsection{Analysis of learned domain weights}

\begin{figure}[h]
  \centering
  \hspace{-1cm}
  \includegraphics[width=1.05\textwidth]{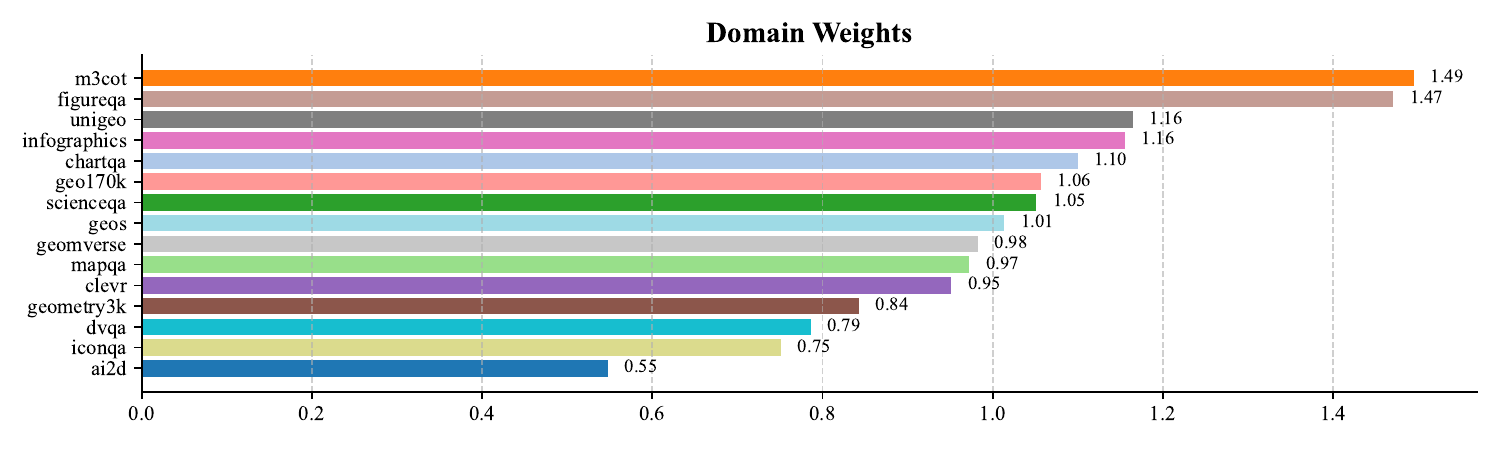} 
  \caption{Learned domain weights after the convergence of the DreamPRM training process.}
  \label{fig:weight_bar}
\end{figure}

The final domain weights (Fig.~\ref{fig:weight_bar}) range from 0.55 to 1.49: \textsc{M3CoT}~\citep{chen2024m3cotnovelbenchmarkmultidomain} and \textsc{FigureQA}~\citep{kahou2018figureqaannotatedfiguredataset} receive the highest weights (approximately 1.5), while \textsc{AI2D}~\citep{kembhavi2016diagramworthdozenimages} and \textsc{IconQA}~\citep{lu2022iconqanewbenchmarkabstract} are assigned lower weights (less than 0.8). This learned weighting pattern contributes to improved PRM performance, indicating that the quality imbalance problem across reasoning datasets is real and consequential. Additionally, as shown in Fig.~\ref{fig:process} in Appendix, all domain weights are initialized to 1.0 and eventually converge during the training process of DreamPRM.



\section{Conclusions}
We propose DreamPRM, the first domain-reweighted PRM framework for multimodal reasoning. By automatically searching for domain weights using a bi-level optimization framework, DreamPRM effectively mitigates issues caused by dataset quality imbalance and significantly enhances the generalizability of multimodal PRMs. Extensive experiments on five diverse benchmarks confirm that DreamPRM outperforms both vanilla PRMs without domain reweighting and PRMs using heuristic data selection methods. We also observe that the domain weights learned by DreamPRM correlate with dataset quality, effectively separating challenging, informative sources from overly simplistic or noisy ones. These results highlight the effectiveness of our proposed automatic domain reweighting strategy.


\section*{Acknowledgments}
This work was supported by the National Science Foundation (IIS2405974 and IIS2339216) and the National Institutes of Health (R35GM157217).

{\footnotesize
\bibliographystyle{plain} 
\bibliography{references}} 





\clearpage
\newpage
\section*{NeurIPS Paper Checklist}

\begin{enumerate}

\item {\bf Claims}
    \item[] Question: Do the main claims made in the abstract and introduction accurately reflect the paper's contributions and scope?
    \item[] Answer: \answerYes{} 
    \item[] Justification: The abstract and introduction faithfully present the contributions and scope of the paper.
    \item[] Guidelines:
    \begin{itemize}
        \item The answer NA means that the abstract and introduction do not include the claims made in the paper.
        \item The abstract and/or introduction should clearly state the claims made, including the contributions made in the paper and important assumptions and limitations. A No or NA answer to this question will not be perceived well by the reviewers. 
        \item The claims made should match theoretical and experimental results, and reflect how much the results can be expected to generalize to other settings. 
        \item It is fine to include aspirational goals as motivation as long as it is clear that these goals are not attained by the paper. 
    \end{itemize}

\item {\bf Limitations}
    \item[] Question: Does the paper discuss the limitations of the work performed by the authors?
    \item[] Answer: \answerYes{} 
    \item[] Justification: We include the limitations of our work in Section~\ref{sec:limitation}.
    \item[] Guidelines:
    \begin{itemize}
        \item The answer NA means that the paper has no limitation while the answer No means that the paper has limitations, but those are not discussed in the paper. 
        \item The authors are encouraged to create a separate "Limitations" section in their paper.
        \item The paper should point out any strong assumptions and how robust the results are to violations of these assumptions (e.g., independence assumptions, noiseless settings, model well-specification, asymptotic approximations only holding locally). The authors should reflect on how these assumptions might be violated in practice and what the implications would be.
        \item The authors should reflect on the scope of the claims made, e.g., if the approach was only tested on a few datasets or with a few runs. In general, empirical results often depend on implicit assumptions, which should be articulated.
        \item The authors should reflect on the factors that influence the performance of the approach. For example, a facial recognition algorithm may perform poorly when image resolution is low or images are taken in low lighting. Or a speech-to-text system might not be used reliably to provide closed captions for online lectures because it fails to handle technical jargon.
        \item The authors should discuss the computational efficiency of the proposed algorithms and how they scale with dataset size.
        \item If applicable, the authors should discuss possible limitations of their approach to address problems of privacy and fairness.
        \item While the authors might fear that complete honesty about limitations might be used by reviewers as grounds for rejection, a worse outcome might be that reviewers discover limitations that aren't acknowledged in the paper. The authors should use their best judgment and recognize that individual actions in favor of transparency play an important role in developing norms that preserve the integrity of the community. Reviewers will be specifically instructed to not penalize honesty concerning limitations.
    \end{itemize}

\item {\bf Theory assumptions and proofs}
    \item[] Question: For each theoretical result, does the paper provide the full set of assumptions and a complete (and correct) proof?
    \item[] Answer: \answerNA{} 
    \item[] Justification: This paper does not include theoretical results.
    \item[] Guidelines:
    \begin{itemize}
        \item The answer NA means that the paper does not include theoretical results. 
        \item All the theorems, formulas, and proofs in the paper should be numbered and cross-referenced.
        \item All assumptions should be clearly stated or referenced in the statement of any theorems.
        \item The proofs can either appear in the main paper or the supplemental material, but if they appear in the supplemental material, the authors are encouraged to provide a short proof sketch to provide intuition. 
        \item Inversely, any informal proof provided in the core of the paper should be complemented by formal proofs provided in appendix or supplemental material.
        \item Theorems and Lemmas that the proof relies upon should be properly referenced. 
    \end{itemize}

    \item {\bf Experimental result reproducibility}
    \item[] Question: Does the paper fully disclose all the information needed to reproduce the main experimental results of the paper to the extent that it affects the main claims and/or conclusions of the paper (regardless of whether the code and data are provided or not)?
    \item[] Answer: \answerYes{} 
    \item[] Justification: All the information needed to reproduce the main experimental results are provided in Section~\ref{sec:prob_set_prelim},~\ref{sec:method}, and~\ref{sec:exp}. We will release the implementation if the paper is accepted.
    \item[] Guidelines:
    \begin{itemize}
        \item The answer NA means that the paper does not include experiments.
        \item If the paper includes experiments, a No answer to this question will not be perceived well by the reviewers: Making the paper reproducible is important, regardless of whether the code and data are provided or not.
        \item If the contribution is a dataset and/or model, the authors should describe the steps taken to make their results reproducible or verifiable. 
        \item Depending on the contribution, reproducibility can be accomplished in various ways. For example, if the contribution is a novel architecture, describing the architecture fully might suffice, or if the contribution is a specific model and empirical evaluation, it may be necessary to either make it possible for others to replicate the model with the same dataset, or provide access to the model. In general. releasing code and data is often one good way to accomplish this, but reproducibility can also be provided via detailed instructions for how to replicate the results, access to a hosted model (e.g., in the case of a large language model), releasing of a model checkpoint, or other means that are appropriate to the research performed.
        \item While NeurIPS does not require releasing code, the conference does require all submissions to provide some reasonable avenue for reproducibility, which may depend on the nature of the contribution. For example
        \begin{enumerate}
            \item If the contribution is primarily a new algorithm, the paper should make it clear how to reproduce that algorithm.
            \item If the contribution is primarily a new model architecture, the paper should describe the architecture clearly and fully.
            \item If the contribution is a new model (e.g., a large language model), then there should either be a way to access this model for reproducing the results or a way to reproduce the model (e.g., with an open-source dataset or instructions for how to construct the dataset).
            \item We recognize that reproducibility may be tricky in some cases, in which case authors are welcome to describe the particular way they provide for reproducibility. In the case of closed-source models, it may be that access to the model is limited in some way (e.g., to registered users), but it should be possible for other researchers to have some path to reproducing or verifying the results.
        \end{enumerate}
    \end{itemize}

\item {\bf Open access to data and code}
    \item[] Question: Does the paper provide open access to the data and code, with sufficient instructions to faithfully reproduce the main experimental results, as described in supplemental material?
    \item[] Answer: \answerYes{} 
    \item[] Justification: We will release the code if the paper is accepted or through an anonymous link per reviewer's request.
    \item[] Guidelines:
    \begin{itemize}
        \item The answer NA means that paper does not include experiments requiring code.
        \item Please see the NeurIPS code and data submission guidelines (\url{https://nips.cc/public/guides/CodeSubmissionPolicy}) for more details.
        \item While we encourage the release of code and data, we understand that this might not be possible, so “No” is an acceptable answer. Papers cannot be rejected simply for not including code, unless this is central to the contribution (e.g., for a new open-source benchmark).
        \item The instructions should contain the exact command and environment needed to run to reproduce the results. See the NeurIPS code and data submission guidelines (\url{https://nips.cc/public/guides/CodeSubmissionPolicy}) for more details.
        \item The authors should provide instructions on data access and preparation, including how to access the raw data, preprocessed data, intermediate data, and generated data, etc.
        \item The authors should provide scripts to reproduce all experimental results for the new proposed method and baselines. If only a subset of experiments are reproducible, they should state which ones are omitted from the script and why.
        \item At submission time, to preserve anonymity, the authors should release anonymized versions (if applicable).
        \item Providing as much information as possible in supplemental material (appended to the paper) is recommended, but including URLs to data and code is permitted.
    \end{itemize}

\item {\bf Experimental setting/details}
    \item[] Question: Does the paper specify all the training and test details (e.g., data splits, hyperparameters, how they were chosen, type of optimizer, etc.) necessary to understand the results?
    \item[] Answer: \answerYes{} 
    \item[] Justification: The detailed experimental settings are included in Section~\ref{sec:expr-setting} and Appendix~\ref{sec:append_1},~\ref{sec:append_2}.
    \item[] Guidelines:
    \begin{itemize}
        \item The answer NA means that the paper does not include experiments.
        \item The experimental setting should be presented in the core of the paper to a level of detail that is necessary to appreciate the results and make sense of them.
        \item The full details can be provided either with the code, in appendix, or as supplemental material.
    \end{itemize}

\item {\bf Experiment statistical significance}
    \item[] Question: Does the paper report error bars suitably and correctly defined or other appropriate information about the statistical significance of the experiments?
    \item[] Answer: \answerNo{} 
    \item[] Justification: Due to the resource limitation, we do not report error bars. But note that we conduct experiments on diverse datasets and follow the protocol used by previous works for fair comparisons.
    \item[] Guidelines:
    \begin{itemize}
        \item The answer NA means that the paper does not include experiments.
        \item The authors should answer "Yes" if the results are accompanied by error bars, confidence intervals, or statistical significance tests, at least for the experiments that support the main claims of the paper.
        \item The factors of variability that the error bars are capturing should be clearly stated (for example, train/test split, initialization, random drawing of some parameter, or overall run with given experimental conditions).
        \item The method for calculating the error bars should be explained (closed form formula, call to a library function, bootstrap, etc.)
        \item The assumptions made should be given (e.g., Normally distributed errors).
        \item It should be clear whether the error bar is the standard deviation or the standard error of the mean.
        \item It is OK to report 1-sigma error bars, but one should state it. The authors should preferably report a 2-sigma error bar than state that they have a 96\% CI, if the hypothesis of Normality of errors is not verified.
        \item For asymmetric distributions, the authors should be careful not to show in tables or figures symmetric error bars that would yield results that are out of range (e.g. negative error rates).
        \item If error bars are reported in tables or plots, The authors should explain in the text how they were calculated and reference the corresponding figures or tables in the text.
    \end{itemize}

\item {\bf Experiments compute resources}
    \item[] Question: For each experiment, does the paper provide sufficient information on the computer resources (type of compute workers, memory, time of execution) needed to reproduce the experiments?
    \item[] Answer: \answerYes{} 
    \item[] Justification: Compute resources used in the experiments are reported in Section~\ref{app:training}.
    \item[] Guidelines:
    \begin{itemize}
        \item The answer NA means that the paper does not include experiments.
        \item The paper should indicate the type of compute workers CPU or GPU, internal cluster, or cloud provider, including relevant memory and storage.
        \item The paper should provide the amount of compute required for each of the individual experimental runs as well as estimate the total compute. 
        \item The paper should disclose whether the full research project required more compute than the experiments reported in the paper (e.g., preliminary or failed experiments that didn't make it into the paper). 
    \end{itemize}
    
\item {\bf Code of ethics}
    \item[] Question: Does the research conducted in the paper conform, in every respect, with the NeurIPS Code of Ethics \url{https://neurips.cc/public/EthicsGuidelines}?
    \item[] Answer: \answerYes{} 
    \item[] Justification: Our paper followed the NeurIPS Code of Ethics.
    \item[] Guidelines:
    \begin{itemize}
        \item The answer NA means that the authors have not reviewed the NeurIPS Code of Ethics.
        \item If the authors answer No, they should explain the special circumstances that require a deviation from the Code of Ethics.
        \item The authors should make sure to preserve anonymity (e.g., if there is a special consideration due to laws or regulations in their jurisdiction).
    \end{itemize}

\item {\bf Broader impacts}
    \item[] Question: Does the paper discuss both potential positive societal impacts and negative societal impacts of the work performed?
    \item[] Answer: \answerYes{} 
    \item[] Justification: Our work helps to enhance multimodal reasoning with DreamPRM. Although the models could still produce errors, we suggest not to rely completely on LLMs and don't perceive it as major negative societal impact.
    \item[] Guidelines:
    \begin{itemize}
        \item The answer NA means that there is no societal impact of the work performed.
        \item If the authors answer NA or No, they should explain why their work has no societal impact or why the paper does not address societal impact.
        \item Examples of negative societal impacts include potential malicious or unintended uses (e.g., disinformation, generating fake profiles, surveillance), fairness considerations (e.g., deployment of technologies that could make decisions that unfairly impact specific groups), privacy considerations, and security considerations.
        \item The conference expects that many papers will be foundational research and not tied to particular applications, let alone deployments. However, if there is a direct path to any negative applications, the authors should point it out. For example, it is legitimate to point out that an improvement in the quality of generative models could be used to generate deepfakes for disinformation. On the other hand, it is not needed to point out that a generic algorithm for optimizing neural networks could enable people to train models that generate Deepfakes faster.
        \item The authors should consider possible harms that could arise when the technology is being used as intended and functioning correctly, harms that could arise when the technology is being used as intended but gives incorrect results, and harms following from (intentional or unintentional) misuse of the technology.
        \item If there are negative societal impacts, the authors could also discuss possible mitigation strategies (e.g., gated release of models, providing defenses in addition to attacks, mechanisms for monitoring misuse, mechanisms to monitor how a system learns from feedback over time, improving the efficiency and accessibility of ML).
    \end{itemize}
    
\item {\bf Safeguards}
    \item[] Question: Does the paper describe safeguards that have been put in place for responsible release of data or models that have a high risk for misuse (e.g., pretrained language models, image generators, or scraped datasets)?
    \item[] Answer: \answerNA{} 
    \item[] Justification: This paper poses no such risks.
    \item[] Guidelines:
    \begin{itemize}
        \item The answer NA means that the paper poses no such risks.
        \item Released models that have a high risk for misuse or dual-use should be released with necessary safeguards to allow for controlled use of the model, for example by requiring that users adhere to usage guidelines or restrictions to access the model or implementing safety filters. 
        \item Datasets that have been scraped from the Internet could pose safety risks. The authors should describe how they avoided releasing unsafe images.
        \item We recognize that providing effective safeguards is challenging, and many papers do not require this, but we encourage authors to take this into account and make a best faith effort.
    \end{itemize}

\item {\bf Licenses for existing assets}
    \item[] Question: Are the creators or original owners of assets (e.g., code, data, models), used in the paper, properly credited and are the license and terms of use explicitly mentioned and properly respected?
    \item[] Answer: \answerYes{} 
    \item[] Justification: We have properly cited papers and models used in our paper.
    \item[] Guidelines:
    \begin{itemize}
        \item The answer NA means that the paper does not use existing assets.
        \item The authors should cite the original paper that produced the code package or dataset.
        \item The authors should state which version of the asset is used and, if possible, include a URL.
        \item The name of the license (e.g., CC-BY 4.0) should be included for each asset.
        \item For scraped data from a particular source (e.g., website), the copyright and terms of service of that source should be provided.
        \item If assets are released, the license, copyright information, and terms of use in the package should be provided. For popular datasets, \url{paperswithcode.com/datasets} has curated licenses for some datasets. Their licensing guide can help determine the license of a dataset.
        \item For existing datasets that are re-packaged, both the original license and the license of the derived asset (if it has changed) should be provided.
        \item If this information is not available online, the authors are encouraged to reach out to the asset's creators.
    \end{itemize}

\item {\bf New assets}
    \item[] Question: Are new assets introduced in the paper well documented and is the documentation provided alongside the assets?
    \item[] Answer: \answerYes{} 
    \item[] Justification: We will release our code with detailed readme files and instructions.
    \item[] Guidelines:
    \begin{itemize}
        \item The answer NA means that the paper does not release new assets.
        \item Researchers should communicate the details of the dataset/code/model as part of their submissions via structured templates. This includes details about training, license, limitations, etc. 
        \item The paper should discuss whether and how consent was obtained from people whose asset is used.
        \item At submission time, remember to anonymize your assets (if applicable). You can either create an anonymized URL or include an anonymized zip file.
    \end{itemize}

\item {\bf Crowdsourcing and research with human subjects}
    \item[] Question: For crowdsourcing experiments and research with human subjects, does the paper include the full text of instructions given to participants and screenshots, if applicable, as well as details about compensation (if any)? 
    \item[] Answer: \answerNA{} 
    \item[] Justification: This work does not involve crowdsourcing nor research with human subjects.
    \item[] Guidelines:
    \begin{itemize}
        \item The answer NA means that the paper does not involve crowdsourcing nor research with human subjects.
        \item Including this information in the supplemental material is fine, but if the main contribution of the paper involves human subjects, then as much detail as possible should be included in the main paper. 
        \item According to the NeurIPS Code of Ethics, workers involved in data collection, curation, or other labor should be paid at least the minimum wage in the country of the data collector. 
    \end{itemize}

\item {\bf Institutional review board (IRB) approvals or equivalent for research with human subjects}
    \item[] Question: Does the paper describe potential risks incurred by study participants, whether such risks were disclosed to the subjects, and whether Institutional Review Board (IRB) approvals (or an equivalent approval/review based on the requirements of your country or institution) were obtained?
    \item[] Answer: \answerNA{} 
    \item[] Justification: This work does not involve crowdsourcing nor research with human subjects.
    \item[] Guidelines:
    \begin{itemize}
        \item The answer NA means that the paper does not involve crowdsourcing nor research with human subjects.
        \item Depending on the country in which research is conducted, IRB approval (or equivalent) may be required for any human subjects research. If you obtained IRB approval, you should clearly state this in the paper. 
        \item We recognize that the procedures for this may vary significantly between institutions and locations, and we expect authors to adhere to the NeurIPS Code of Ethics and the guidelines for their institution. 
        \item For initial submissions, do not include any information that would break anonymity (if applicable), such as the institution conducting the review.
    \end{itemize}

\item {\bf Declaration of LLM usage}
    \item[] Question: Does the paper describe the usage of LLMs if it is an important, original, or non-standard component of the core methods in this research? Note that if the LLM is used only for writing, editing, or formatting purposes and does not impact the core methodology, scientific rigorousness, or originality of the research, declaration is not required.
    \item[] Answer: \answerYes{} 
    \item[] Justification: LLMs, specifically MLLMs, are used in the experiments as the paper is about multimodal reasoning. The usage is described in Secion~\ref{sec:prob_setting},~\ref{sec:method}. In terms of writing, LLMs are only used for checking grammar, spelling, and word choices.
    \item[] Guidelines:
    \begin{itemize}
        \item The answer NA means that the core method development in this research does not involve LLMs as any important, original, or non-standard components.
        \item Please refer to our LLM policy (\url{https://neurips.cc/Conferences/2025/LLM}) for what should or should not be described.
    \end{itemize}

\end{enumerate}

\clearpage

\appendix
\section*{Appendix}
\section{Optimization algorithm}
\label{app:optim-algo}
Directly solving the bi-level optimization problem in Equation \ref{eq:upper} can be computational prohibitive due to its nested structure. Following previous work~\citep{choe2023betty}, we use approximated algorithm with a few unrolling steps. For example, under one-step unrolling, the updating of PRM's weights can be expressed as:


\begin{equation}
\label{eq:optim-lower}
\phi^{(t+1)} = \phi^{(t)} - \beta_1  \nabla_{\phi} \mathcal{L}_{tr}(\mathcal D_{tr}, \phi , \alpha)
\end{equation}

where $\beta_1$ is the learning rate in lower level optimization. After obtaining the updated PRM parameter  $\phi^{(t+1)}$ from Equation \ref{eq:optim-lower}, the domain-reweighting parameter $\alpha$ is then updated as follows:



\begin{equation}
\label{eq:optim-upper}
\alpha^{(t+1)} = \alpha^{(t)} - \beta_2 \nabla_{\alpha} \mathcal{L}_{meta}(\mathcal D_{meta}, \phi^*(\alpha))
\end{equation}

where $\beta_2$ is the learning rate for upper level optimization. The two optimization steps in Equation \ref{eq:optim-lower} and Equation \ref{eq:optim-upper} are conducted iteratively until convergence to get optimal PRM weights $\phi^*$ and optimal domain reweighting parameter $\alpha^*$.

\section{Datasets and benchmarks} 
\label{sec:append_1}

\begin{table}[h]
\label{tab:data}
\centering
\rowcolors{2}{gray!15}{white}
\caption{Multimodal datasets involved in the fine-tuning of DreamPRM, organized by task category.}
\begin{tabular}{l p{10cm}}
\toprule
\rowcolor{white}   
\textbf{Task} & \textbf{Dataset} \\
\midrule
Science     & AI2D~\citep{kembhavi2016diagramworthdozenimages}, ScienceQA~\citep{lu2022learnexplainmultimodalreasoning}, M3CoT~\citep{chen2024m3cotnovelbenchmarkmultidomain} \\
Chart       & ChartQA~\citep{masry2022chartqabenchmarkquestionanswering}, DVQA~\citep{kafle2018dvqaunderstandingdatavisualizations}, MapQA~\citep{chang2022mapqadatasetquestionanswering}, FigureQA~\citep{kahou2018figureqaannotatedfiguredataset} \\
Geometry    & Geo170k~\citep{gao2023gllavasolvinggeometricproblem}, Geometry3K~\citep{lu2021inter}, UniGeo~\citep{chen2022unigeounifyinggeometrylogical}, GeomVerse~\citep{kazemi2023geomversesystematicevaluationlarge}, GeoS~\citep{seo-etal-2015-solving} \\
Commonsense & IconQA~\citep{lu2022iconqanewbenchmarkabstract}, InfographicsVQA~\citep{mathew2021infographicvqa}, CLEVR-Math~\citep{lindstrom2022clevrmathdatasetcompositionallanguage} \\
\bottomrule
\end{tabular}
\label{tab:vqa_datasets}
\end{table}

For datasets used in lower-level optimization ($\mathcal{D}_{tr}$ in Section \ref{sec:method}), our study utilizes a diverse set of datasets, spanning multiple domains to ensure a comprehensive coverage of multimodal reasoning tasks, as reported in Tab.~\ref{tab:vqa_datasets}. The selected 15 multimodal datasets covers 4 major categories including science, chart, geometry and commonsense, with a wide range of task types (QA, OCR, spatial understanding). 
Additionally, we observe that for some questions, given the current structural thinking prompts, MLLMs consistently produce either correct or incorrect answers. Continuing to sample such questions is a waste of computational resources. Inspired by the dynamic sampling strategy in DAPO~\citep{yu2024mmvetevaluatinglargemultimodal}, we propose a similar dynamic sampling technique for Monte Carlo estimation that focuses on prompts with varied outcomes to improve efficiency. After processing and sampling, the training datasets in lower-level $\mathcal D_{tr}$ have around 15k examples (1k per each of the 15 domains), while the meta dataset in the upper-level  $\mathcal D_{meta}$ has around 1k validation examples from the MMMU~\citep{yue2024mmmumassivemultidisciplinemultimodal} dataset.

For the dataset used in upper-level optimization ($\mathcal{D}_{meta}$ in Section \ref{sec:method}), we select data from \textsc{MMMU}~\citep{yue2024mmmumassivemultidisciplinemultimodal} to simulate a realistic and diverse reasoning scenario. \textsc{MMMU} focuses on advanced perception and reasoning with domain-specific knowledge. Its questions span 30 subjects and 183 subfields, comprising 30 highly heterogeneous image types, such as charts, diagrams, maps, tables, music sheets, and chemical structures.

At evaluation time, we use five multimodal reasoning benchmarks for testing the capability of DreamPRM. \textsc{WeMath}~\citep{qiao2024wemathdoeslargemultimodal}, \textsc{MathVista}~\citep{lu2024mathvista}, and \textsc{MathVision}~\citep{wang2024measuringmultimodalmathematicalreasoning} focus more on math-related reasoning tasks and logic and critical thinking, while \textsc{MMVet}~\citep{yu2024mmvetevaluatinglargemultimodal} and \textsc{MMStar}~\citep{chen2024rightwayevaluatinglarge} focus more on real-life tasks that require common knowledge and general reasoning abilities.

\section{Structural Thinking Prompt}\label{app:st}
\begin{figure}[h]
  \centering
  \includegraphics[width=\textwidth]{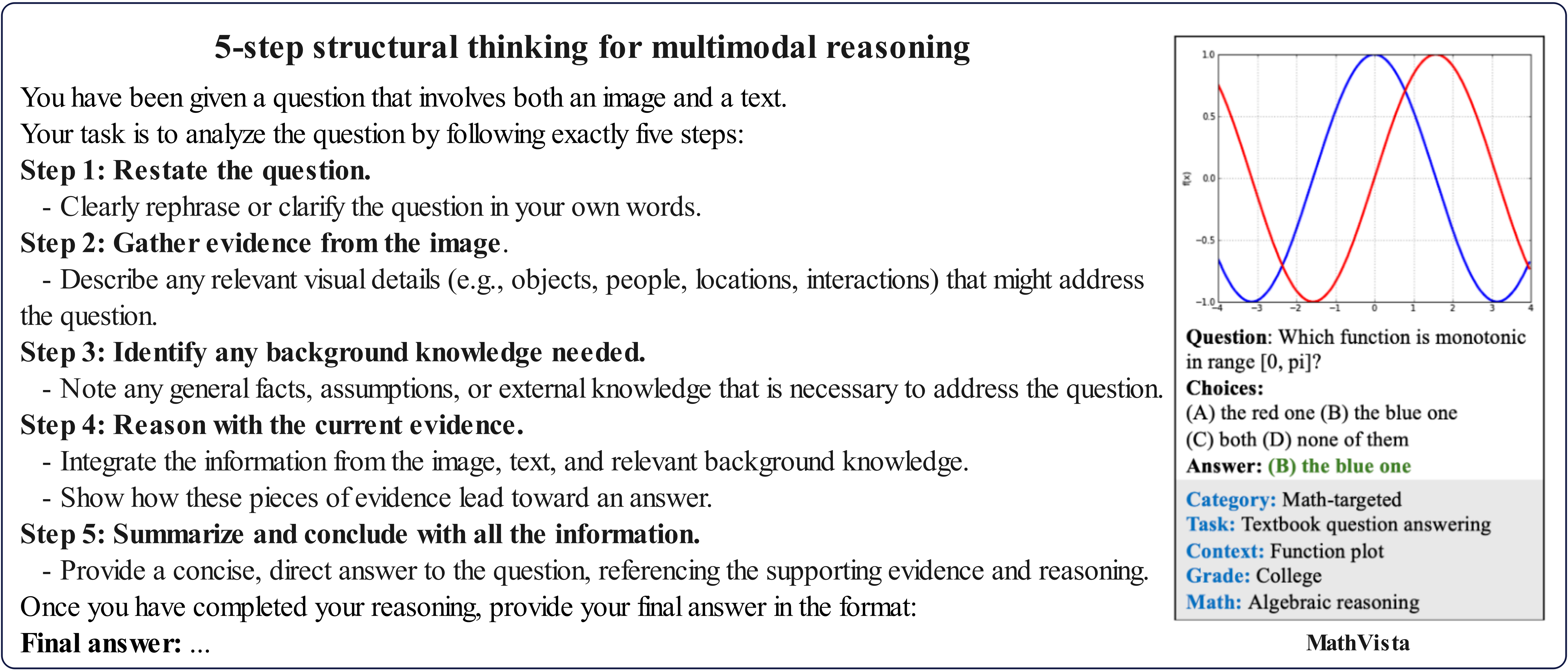} 
  \caption{Zero-shot prompting for structural thinking.}
  \label{fig:structural thinking}
\end{figure}

The detailed structural thinking prompt applied in our experiments is reported in Fig.~\ref{fig:structural thinking}. We carefully design 5 reasoning steps to boost the reasoning capabilities of the MLLMs and enable process supervision.
\label{sec:append_2}

\begin{table}[htbp]
\caption{Accuracy on MathVista using DreamPRM with varying numbers $k$ of CoTs.}
\centering
\begin{tabular}{lccccc}
\toprule
\multirow{2}{*}{\textbf{Model Name}} 
& \multicolumn{1}{c}{pass@1} 
& \multicolumn{4}{c}{\textbf{DreamPRM (select $k$ CoTs)}} \\ \cmidrule(lr){2-2}\cmidrule(lr){3-6}
& \textbf{$k{=}1$} & $k{=}2$ & $k{=}4$ & $k{=}6$ & $k{=}8$ \\
\midrule
InternVL-2.5-8B-MPO~\citep{wang2024mpo}               & 65.4 & 65.3 & 66.5 & 67.8 & \bf68.9 \\
GPT-4.1-mini (4-14-25)~\citep{openai2024gpt4technicalreport} & 71.5 & 71.8 & 72.5 & 73.2 & \bf74.4 \\
\bottomrule
\end{tabular}
\label{tab:bestofN}
\end{table}

\begin{table}[htbp]
\caption{Ablation study evaluating the impact of individual components of DreamPRM}
\centering
\resizebox{\textwidth}{!}{
\begin{tabular}{lccccc}
\toprule
\textbf{Ablation / Dataset} & \textbf{WeMath} & \textbf{MathVista} & \textbf{MathVision} & \textbf{MMVet} & \textbf{MMStar} \\
\midrule
\textbf{DreamPRM (original)} &\bf 57.4&\bf 68.9&\bf 22.1&\bf 61.4&\bf62.3\\
\quad w/o aggregation function loss        & 56.3 (-1.1)   & 66.1 (-2.8)   & 20.1 (-2.0)   & 60.0 (-1.4)   & 59.6 (-2.7)\\
\quad w/o bi-level optimization     & 55.0 (-2.4) & 65.4 (-3.5) & 19.9 (-2.2) & 61.2 (-0.2) & 58.9 (-3.4) \\
\quad w/o structural thinking       & 54.6 (-2.8) & 65.7 (-3.2) & 20.3 (-1.8) & 57.5 (-3.9) & 61.6 (-0.7)\\

\bottomrule
\end{tabular}}
\label{tab:ablation}
\end{table}

\section{Additional Experimental Results}\label{app:additional}
\textbf{Leaderboard performance details.} Table~\ref{tab:prm_comparison} presents a comprehensive comparison of different PRM variants built upon the same o4-mini backbone.
DreamPRM consistently outperforms all baselines, elevating the base o4-mini performance from 80.6
These steady improvements demonstrate the effectiveness of DreamPRM in enhancing reasoning accuracy through process-level supervision and promoting more reliable consensus across multiple chains of thought.

\textbf{Best-of-N results.} Tab.~\ref{tab:bestofN} reports the accuracy of two state-of-the-art models on MathVista dataset using DreamPRM with varying numbers $k$ of CoTs. The results indicate that the performance scales well with the number of CoTs.

\begin{table}[t]
\centering
\renewcommand{\arraystretch}{1.1}
\setlength{\tabcolsep}{8pt}
\caption{
Comparison of different PRM variants on the o4-mini model (evaluated on eight CoTs).\newline
}
\label{tab:prm_comparison}
\begin{tabular}{lc}
\toprule
\textbf{Method} & \textbf{Accuracy} \\
\midrule
o4-mini                         & 80.6 \\
 \quad+ Self-consistency      & 81.7 \\
 \quad+ ORM                   & 80.8 \\
 \quad+ Vanilla-PRM           & 84.2 \\
 \quad+ DreamPRM              & \textbf{85.2} \\
\bottomrule
\end{tabular}
\end{table}

\textbf{Ablation studies.} The exact results of ablation experiments in the main paper are included in Tab.~\ref{tab:ablation}, which emphasizes the importance of all the components in DreamPRM.

\begin{figure}[t]
  \centering
  \includegraphics[width=\textwidth]{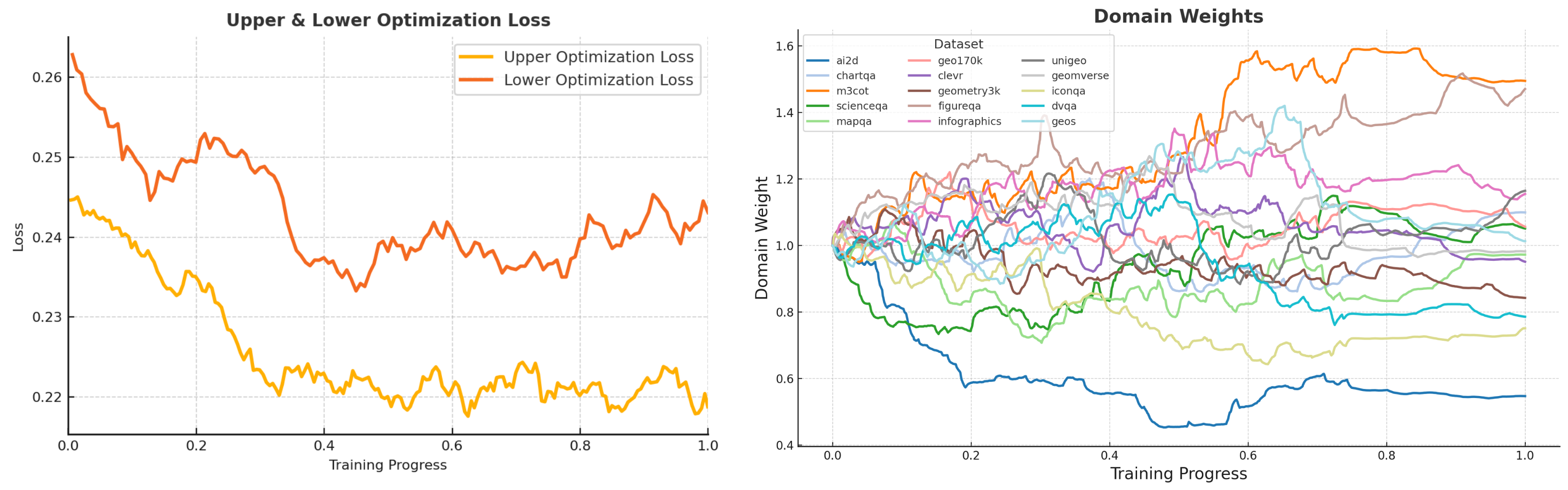} 
  \caption{Optimization loss curves and dynamic domain weights throughout DreamPRM fine-tuning.}
  \label{fig:process}
\end{figure}

\textbf{Loss curves and domain weights.} The loss curves and domain weights during the fine-tuning of DreamPRM are illustrated in Fig.~\ref{fig:process}. It can be observed that the learnt distribution emphasizes informative mathematical figure domains while attenuating less relevant sources. Additionally, domain weights start at 1.0 and quickly diverge, stabilizing after roughly half the training, and the inner and outer losses decrease steadily and plateau, indicating stable convergence of the bi‑level training procedure. 

\begin{figure}[h]
  \centering
  \includegraphics[width=\textwidth]{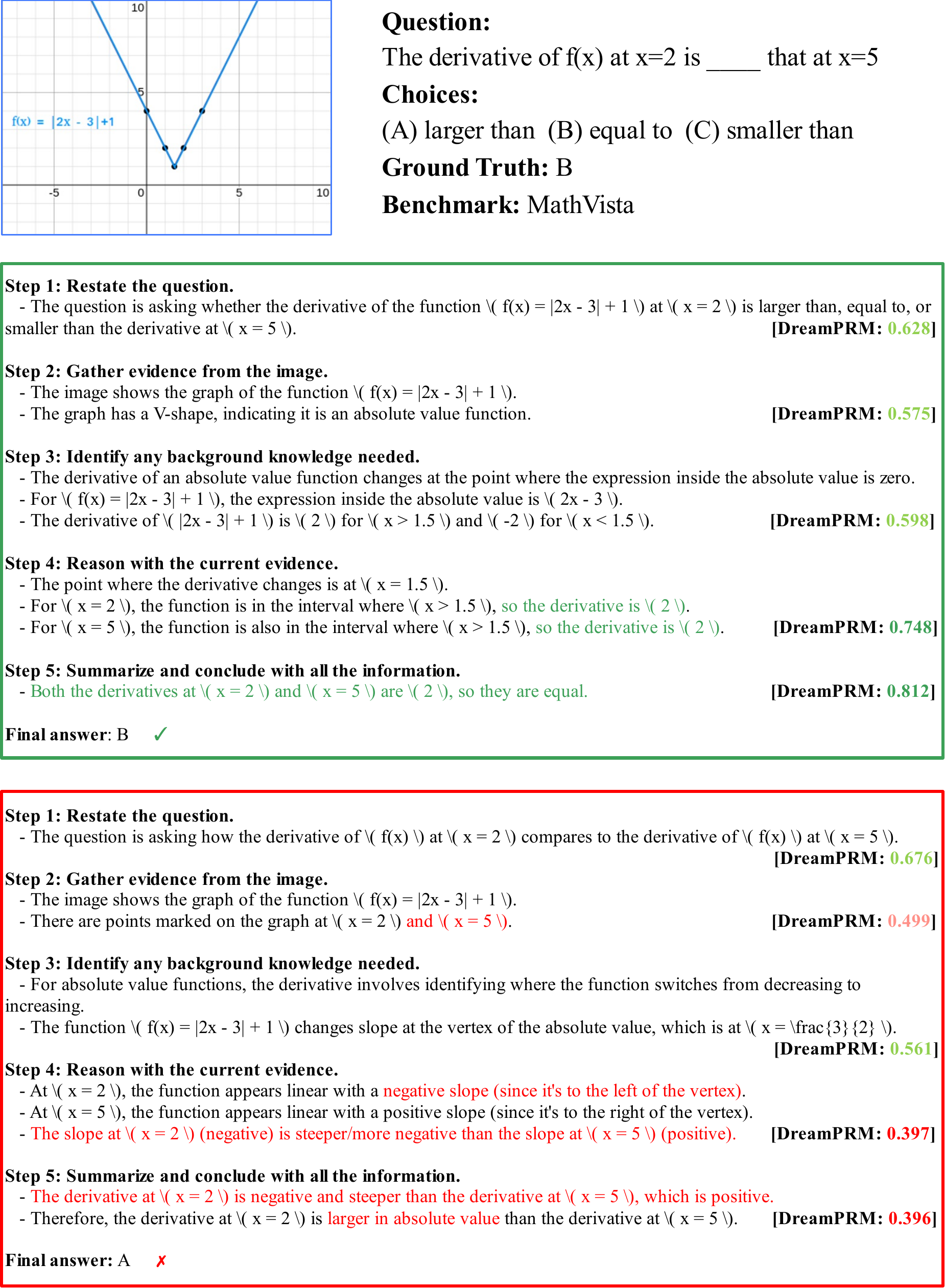} 
  \caption{A case study of DreamPRM's step-wise evaluation.}
  \label{fig:case}
\end{figure}

\textbf{Case study.} A complete case study illustrating DreamPRM's step-wise evaluation is reported in Fig.~\ref{fig:case}. DreamPRM assigns higher scores to high-quality, coherent reasoning steps, while penalizes flawed or unsupported steps. 

\section{Limitations \& Future Work.}
\label{sec:limitation}
DreamPRM currently assumes a fixed set of domains and requires Monte-Carlo sampling, which can be computationally heavy.  
Future work could explore instance-level reweighting, adaptive sampling strategies, and integration with retrieval-augmented generation to further cut compute while broadening coverage. We will release code, trained weights, and evaluation scripts to facilitate reproducibility and community adoption.


\end{document}